# Searching the Title of Practical Work of the Informatics Engineering Bachelor Program with the Case Base Reasoning Method

Agung Sukrisna Jaya [a,1,*], Osvari Arsalan [b,2], Danny Matthew Saputra [b,3]

[a] Department of Informatics, Sriwijaya University, Palembang, Indonesia
[b] Department of Informatics, Faculty of Computer Science, Sriwijaya University, Palembang, Indonesia
[1] 09021281823045@student.unsri.ac.id; [2] osvari.arsalan@ilkom.unsri.ac.id; [3] danny@ilkom.unsri.ac.id
* corresponding author

ABSTRACT
Case-Based Reasoning (CBR) is a problem-solving approach that utilizes past cases to address new problems by identifying the highest similarity between them. This study applies CBR to the retrieval of practical work titles. Term Frequency–Inverse Document Frequency (TF-IDF) is used for vectorizing title words, and Cosine Similarity is employed to calculate similarity values. The system supports searches using complete titles or keywords and provides relevant titles along with their similarity scores. Testing was conducted using a dataset of 705 practical work titles, with five titles queried in two stages: the first using existing titles and the second using randomized titles from the first stage. The results indicate that the second stage produced the same number of retrieved titles and maintained the highest average similarity score.

## 1. Introduction

The advancement of technology and information has led to a rapid growth in various fields. Undoubtedly, the global community extensively relies on technology as a solution to address the myriad challenges of the contemporary world. One prominent application is the search systems, which offer efficient methods for locating specific information within vast data collections. For instance, a search system can be employed to locate titles of student practical work [1].

A search engine is the practical application of information retrieval techniques for large the term "search engine" was originally term "Search Engine" was originally used to refer to specialized hardware for text searching [2].

Among the problem-solving techniques rooted in historical knowledge, Case-Based Reasoning stands out. This approach leverages past experiences as a foundation for addressing current challenges. The repository where past instances are stored is known as the "case base." In this repository, previous scenarios are cataloged, serving as a valuable resource to draw upon when tackling analogous situations in the present and future [3].

## 2. Literature Study

This chapter focusing literature review as the foundation this research, the main discussion is Searching, Case Base Reasoning, TF-IDF, Cosine Similarity Algorithm.

*A. Searching*

Searching is a method to find information/data that is being searched in a data set that has the same data type. Search is needed to get information that is either unknown or already known. There two types of searching Sequential Search and Binary Search, Sequential Search is a method of finding data from data set where data is searched from front to back or from the beginning to the end of data without having to sort the data [4]. Binary Search is a method of finding data from data set where the data set must be in the correct order so that data search process can be carried out [5].

## B. Case Base Reasoning

Case Base Reasoning is a method for reasoning or solving problem based on exisring case to solve new problems [6]. Case Base Reasoning has four cycles is retrieve, reuse, revise and retain. Retrive is to get existing cases that are similar to the new case. Reuse is using similar previous cases and used to solve the current problem, Revise is adapting previous solutions to solve current problems. Retain is integrate the new solution so that can be used by similar future cases [7]. An overview cycles of Case Base Reasoning can be seen in Fig 1.

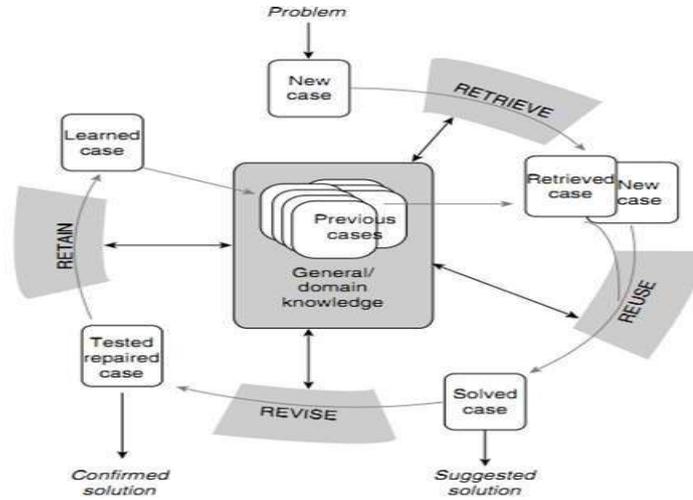

Fig. 1. Cycles of Case Base Reasoning (Supriana & Dwi Prebiana, n.d.)

## C. TF-IDF

TF-IDF is a type of weighting that involves statistical measurements to measure how important a word is in the document set [8]. TF-IDF is an integration method between Term Frequecy (TF) and Inverse Document Frequency (IDF). Term Frequency is calculated using the i-th equation (1). The i-th Term Frequency is the occurrence of the i-th Term in the j-th document. Inverse Document Frequecy (IDF) algorithm from the ratio of the total number of documents in corpus to the number of documents that have a Term in equation (2). The value obtained is multiplied by both in equation (3). The TF-IDF function is used to find the value of each document from a training data set where the data will be formed into a vector between documents and words. [9].

$$tf_i \frac{freq_i(d_j)}{\sum_{i=1}^{k} freq_i(d_j)} \quad (1)$$

$$idf_i = \log \frac{|D|}{|\{d : t_i \in d\}|} \quad (2)$$

$$(tf - idf)_{ij} = tf_i(d_i) * idf_i \quad (3)$$

Description :

$tf_i$ = The i-th Term Frequency

$freq_i$ = Frequency i

$d_i$ = i-th document

$d_j$ = j-th document

$\sum^{k}_{i=1}$ = sigma from 1 to k $idf_i$ = Inverse Document Frequency i

$t_i$ = Term i

$D$ = Corpus of Documents

*D. Cosine Similarity Algorithm*

Cosine Similarity is an algorithm used to calculate how much similarity between documents, Cosine Similarity uses a similarity measure function. Using a measure function allows ranking documents according to their similarity or relevance to the query [10].

There few steps in Cosine Similarity Algorithm, Calculating the vector similarity of the query with each existing document. The calculation is done by calculating the result of the multiplication between the query vector and other documents, then calculating the length of each document, including the query vector by squaring the weight of each term for each document, implementing the calculation of the Cosine Similarity formula calculating the similarity of the query vector with the existing document and sort the calculation result from the largest [11].

description :

$$Similarity(X, Y) = \frac{|X \cap Y|}{|X|^{1/2}|Y|^{1/2}} \quad (4)$$

$|X \cap Y|$ = Number of terms contained in document X and document Y

$|X|$ = Number of terms contained in document X

$|Y|$ = Number of terms contained in document Y

## 3. Methodology

*A. Data Collection*

The type of data used is 705 secondary data, the data used is the title of practical work obtained from the administration of the informatics engineering bachelor program.

*B. Framework*

Framework of Searching system built in this research can be seen in Fig. 2.

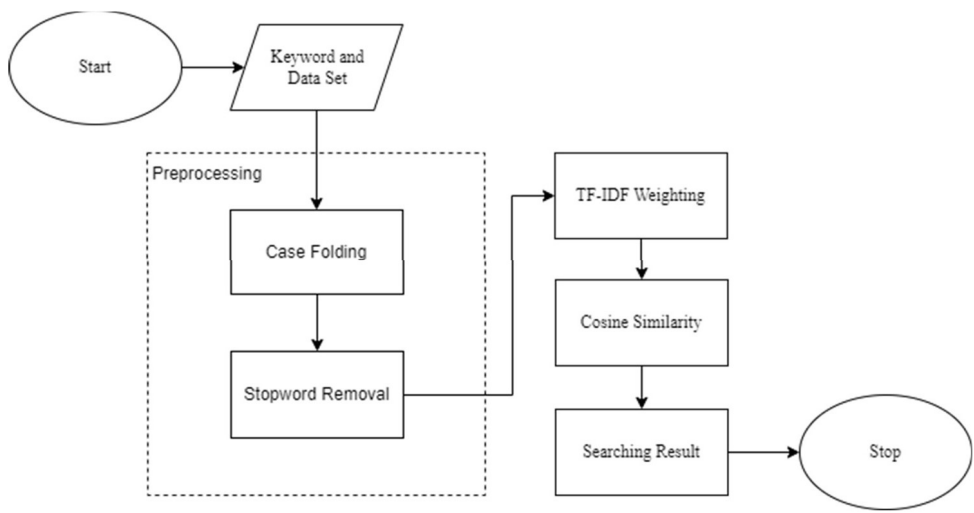

Fig. 2. Searching System Framework

Process building The Searching system based on Fig. 3, system will go through several stages. First, the keywords and dataset will be preprocessed.

Next stage, the data set will get weighted by TF-IDF. Then, the keyword will be converted into the same value as the document that has been processed by TF-IDF by Cosine Similarity and search for relevant practical work titles with keywords.

## 4. Result and Discussion

Testing is carried out in two stages, first the system performs a search based on the keywords entered by the user and sees how many practical work titles are found by the system, the second stage of the system searches for the first stage keywords that have randomized the position of each word and sees the number of practical work titles found and the highest match score with the first stage keywords.

**Table 1.** First Stage Testing

| Keyword | Title Found |
| --- | --- |
| Sistem Pendukung Keputusan Promosi dan Evaluasi Kinerja Karyawan | 371 Title |
| Apliaksi Reminder Pembayaran Tagihan Flexi Home | 254 Title |
| Sistem Navigasi Gedung dengan Metode Algoritma Djikstra | 378 Title |
| Sistem Administrasi Realisasi Kredit BRIGUNA | 367 Title |
| Perancangan dan Implementasi Aplikasi Sistem Monitoring | 600 Title |

**Table 2.** Second Stage Testing

| Keyword | Title Found | Highest Match Score |
| --- | --- | --- |
| Pendukung Sistem Promosi Jabatan Keputusan dan Kinerja Evaluasi Karyawan | 371 Title | 1.0 |
| Reminder Aplikasi Tagihan Pembayaran Home Flexi | 254 Title | 1.0 |
| Navigasi Sistem Gedung dengan Algoritma Djikstra Metode | 378 Title | 1.0 |
| Realisasi Administrasi Sistem Kredit Briguna | 367 Title | 1.0 |
| Implementasi dan Perancangan Sistem Surat Monitoring Aplikasi | 600 Title | 1.0 |

In table 1, the first stage of testing has been carried out using five titles in the document and resulted in many practical work titles found by the system. In table 2, the media stage testing has been carried out using five titles in the first stage testing, each of which has been randomized.

Based on the results of testing that has been carried out through two stages, the results of research that has been carried out through two stages of testing with five titles, it is found that the second stage of testing produces many of the same titles as the first stage of testing and each title in the second stage has the highest average match score of 1.0.

## 5. Conclusion

The results of the search for practical work titles with the case base reasoning method that has been carried out through two stages of testing with five titles, it is found that the second stage of testing produces many of the same titles as the first stage of testing and each title in the second stage has the highest average match score of 1.0.